\documentclass{article}
\usepackage{times}
\usepackage{graphicx}
\usepackage{natbib}
\usepackage{algorithm}
\usepackage{algorithmic}

\AtBeginDocument{%
 \abovedisplayskip=4pt plus 0pt minus 0pt
 \abovedisplayshortskip=1pt plus 0pt
 \belowdisplayskip=4pt plus 0pt minus 0pt
 \belowdisplayshortskip=1pt plus 0pt minus 0pt
}
\usepackage{hyperref}

\usepackage{booktabs}
\usepackage{blindtext}
\usepackage{graphicx}
\usepackage{caption,subcaption}
\usepackage{etoolbox}
\usepackage{bm}
\usepackage{scrextend}
\usepackage[flushleft]{threeparttable}
\usepackage[usenames,dvipsnames,svgnames,table]{xcolor}
\usepackage[normalem]{ulem}
\usepackage{amsmath,amsthm, amssymb, latexsym}
\usepackage{epsfig}
\usepackage{multirow}
\usepackage{pifont}
\usepackage{bigints}
\usepackage{soul}
\usepackage{marginnote}
\usepackage{colortbl}
\usepackage{subcaption}

\newcommand{\var}{{\rm I\kern-.3em D}}

\usepackage[accepted]{icml2017} 
\icmltitlerunning{Bayesian Sparsification for Recurrent Neural Networks}

\begin{document} 

\twocolumn[
\icmltitle{Bayesian Sparsification of Recurrent Neural Networks}



% you can specify symbols, otherwise they are numbered in order
\icmlsetsymbol{equal}{*}

\begin{icmlauthorlist}
\icmlauthor{Ekaterina Lobacheva}{equal,hse,kl}
\icmlauthor{Nadezhda Chirkova}{equal,hse,msu}
\icmlauthor{Dmitry Vetrov}{hse,yandex}
\end{icmlauthorlist}

\icmlaffiliation{hse}{National Research University Higher School of Economics, Moscow, Russia}
\icmlaffiliation{msu}{Lomonosov Moscow State University, Moscow, Russia}
\icmlaffiliation{kl}{Kaspersky Lab, Moscow, Russia}
\icmlaffiliation{yandex}{Yandex, Moscow, Russia}

\icmlcorrespondingauthor{Ekaterina Lobacheva}{elobacheva@hse.ru}
\icmlcorrespondingauthor{Nadezhda Chirkova}{nchirkova@hse.ru}
%\icmlcorrespondingauthor{Dmitry Vetrov}{dvetrov@hse.ru}

\icmlkeywords{variational dropout, recurrent neural networks, sparsification, regularization}

\vskip 0.3in
]



\printAffiliationsAndNotice{\icmlEqualContribution} % otherwise use the standard text.

\begin{abstract} 
Recurrent neural networks show state-of-the-art results in many text analysis tasks but often require a lot of memory to store their weights. Recently proposed Sparse Variational Dropout~\cite{dmolch} eliminates the majority of the weights in a feed-forward neural network without significant loss of quality. We apply this technique to sparsify recurrent neural networks. To account for recurrent specifics we also rely on Binary Variational Dropout for RNN~\cite{gal}. We report 99.5\% sparsity level on sentiment analysis task without a quality drop and up to 87\% sparsity level on language modeling task with slight loss of accuracy.
\end{abstract} 

\section{Introduction}
\label{intro}
Recurrent neural networks (RNNs) are among the most powerful models for natural language processing, speech recognition, question-answering systems and other problems with sequential data~\cite{las,deepspeech,hypernet,trans,qa}. For complex tasks such as machine translation~\cite{trans} or speech recognition~\cite{deepspeech} modern RNN architectures incorporate a huge number of parameters. To use these models on portable devices with limited memory, for instance, smartphones, the model compression is desired. High compression level may also lead to an acceleration of RNNs. In addition, compression regularizes RNNs and helps to avoid overfitting.

There are a lot of RNNs compression methods based on specific weight matrix representations~\cite{tjandra,kroneker} or sparsification via pruning~\cite{pruning}. In this paper we focus on RNNs compression via sparsification. Most of the methods from this group are heuristic and require time-consuming hyperparameters tuning.

Recently Molchanov et. al.~\yrcite{dmolch} proposed a principled method based on variational dropout for sparsification of fully connected and convolutional networks. A probabilistic model was described in which parameters controlling sparsity are tuned automatically during neural network training. This model called Sparse Variational Dropout (Sparse VD) leads to extremely sparse solutions without a significant quality drop. However, this technique was not previously investigated for RNNs.

In this paper we apply Sparse VD to recurrent neural networks. To take into account the specifics of RNNs we rely on some insights underlined in the paper by Gal \& Ghahramani~\yrcite{gal} where they explain the proper way to use binary dropout in RNNs from the Bayesian point of view. In the experiments we show that LSTMs with Sparse VD yield high sparsity level with just a slight drop in quality. We achieved 99.5\% sparsity level on sentiment analysis task and up to 87.6\% in character level language modeling experiment.

\section{Preliminaries}
\label{preliminaries}

\subsection{Bayesian Neural Networks}
Consider a neural network with weights $\omega$ modeling the dependency of the target variables $y=\{y^1, \dots, y^\ell\}$ on the corresponding input objects $X = \{x^1, \dots, x^\ell\}$.
In a Bayesian neural network the weights $\omega$ are treated as random variables. With the prior distribution $p(\omega)$ we search for the posterior distribution $p(\omega|X, y)$ that will help to find expected target value during inference.  
In the case of neural networks, true posterior is usually intractable but it can be approximated by some parametric distribution $q_\lambda(\omega)$. 
The quality of this approximation is measured by the KL-divergence $KL(q_\lambda(\omega)||p(\omega|X, y))$.
The optimal parameter $\lambda$ can be found by maximization of the variational lower bound w.r.t. $\lambda$:
\begin{equation}
\label{VLB}
\mathcal{L}=\sum_{i=1}^\ell \mathbb{E}_{q_\lambda(\omega)} \log p(y^i|x^i, \omega) - KL(q_\lambda(\omega)||p(\omega))
\end{equation}
The expected log-likelihood term in \eqref{VLB}
is usually approximated by Monte-Carlo sampling.
To make the MC estimation unbiased, the weights are parametrized by a deterministic function: $\omega = g(\lambda, \xi)$, where $\xi$ is sampled from some non-parametric distribution (the reparameterization trick~\cite{rt}). 
The KL-divergence term in \eqref{VLB}
acts as a regularizer and is usually computed or approximated analytically.

\subsection{Sparse Variational Dropout}
Dropout~\cite{dropout} is a standard technique for regularization of neural networks. It implies that inputs of each layer are multiplied by a randomly generated noise vector. 
The elements of this vector are usually sampled from Bernoulli or Gaussian distribution with the parameters tuned using cross-validation. Kingma et al.~\yrcite{kingma} interpreted Gaussian dropout from a Bayesian perspective that allowed to tune dropout rate automatically during model training. Later this model was extended to sparsify fully connected and convolutional neural networks resulting in a model called Sparse Variational Dropout (Sparse VD)~\cite{dmolch}.

Consider one dense layer of a feed-forward neural network with an
input of the size $n$, an output of the size $m$ and 
a weight matrix $W$. 
Following Kingma et al.~\yrcite{kingma}, in Sparse VD the prior on the weights is a fully factorized log-uniform distribution $p(|w_{ij}|) \propto \frac 1 {|w_{ij}|}$ and the posterior is searched in the form of fully factorized normal distribution: 
\begin{equation}
q(w_{ij}|m_{ij}, \alpha_{ij}) = \mathcal{N}(m_{ij}, \alpha_{ij} m^2_{ij}).
\end{equation}
Employment of such form of the posterior distribution is equivalent to putting multiplicative~\cite{kingma} or additive~\cite{dmolch} normal noise on the weights in the following manner:
\begin{equation}
w_{ij} = m_{ij} \xi_{ij}, \quad \xi_{ij}\sim\mathcal{N}(1, \alpha_{ij}),
\end{equation}
%\vspace{-0.3cm}
\begin{equation}
\label{ART}
w_{ij} = m_{ij} + \epsilon_{ij}, \quad \epsilon_{ij}\sim\mathcal{N}(0, \sigma^2_{ij}), \quad \alpha_{ij} = \frac{\sigma^2_{ij}}{m^2_{ij}}.
\end{equation}
The representation~\eqref{ART} is called additive reparameterization~\cite{dmolch}. It reduces the variance of the gradients of $\mathcal{L}$ w.\,r.\,t. $m_{ij}$. Moreover, since a sum of normal distributions is a normal distribution with computable parameters, the noise may be applied to the preactivation (input vector times weight matrix $W$) instead of $W$. This trick is called the local reparameterization trick~\cite{fast_dropout,kingma} and it reduces the variance of the gradients even further and makes training more efficient.

In Sparse VD optimization of the variational lower bound~\eqref{VLB} is performed w.\,r.\,t. $\{M, \log \sigma\}$. The KL-divergence factorizes over the weights and its terms depend only on $\alpha_{ij}$ because of the specific choice of the prior~\cite{kingma}:
\begin{equation}
KL(q(w_{ij}|m_{ij}, \alpha_{ij})||p(w_{ij}))=k(\alpha_{ij}).
\end{equation}
Each term can be approximated as follows~\cite{dmolch}:
\begin{equation}
\begin{gathered}
k(\alpha) \approx 0.64 \sigma (1.87 + 1.49\log \alpha)-\\
\:\:\:\,- 0.5 \log(1 + \alpha^{-1}) + C.
\label{kl}
\end{gathered}
\end{equation}

KL-divergence term encourages large values of $\alpha_{ij}$. If $\alpha_{ij} \rightarrow \infty$ for a weight $w_{ij}$, the posterior over this weight is a high-variance normal distribution and it is beneficial for model to put   $m_{ij} = 0$ as well as $\sigma_{ij}=\alpha_{ij} m^2_{ij}=0$ to avoid inaccurate predictions. As a result, the posterior over $w_{ij}$ approaches zero-centered $\delta$-function, the weight does not affect the network's output and can be ignored. 

\subsection{Dropout for Recurrent Neural Networks}
Yet another Bayesian model was proposed by Gal \& Ghahramani~\yrcite{bindrop} to explain the binary dropout. On this base, a recipe how to apply a binary dropout to the RNNs properly was proposed by Gal \& Ghahramani~\yrcite{gal}.

The recurrent neural network takes a sequence $x = [x_0, \dots, x_T]$, $x_t\in \mathbb{R}^n$ as an input and maps it into the sequence of hidden states:
\begin{equation}
\begin{gathered}
h_{t} = f_h(x_t, h_{t-1}) = g_h(x_{t} W^x  + h_{t-1} W^h  + b_1)\\
h_i \in \mathbb{R}^m, \, h_0 = \bar 0
\end{gathered}
\end{equation}

Throughout the paper, we assume that the output of the RNN depends only on the last hidden state:
\begin{equation}
y = f_y(h_T) = g_y(h_T W^y + b_2).
\end{equation}
Here $g_h$ and $g_y$ are some nonlinear functions.
However, all the techniques we discuss further can be easily applied to the more complex setting, e.\,g. language model with several outputs for one input sequence (one output for each time step).

Gal \& Ghahramani~\yrcite{gal} considered RNNs as Bayesian networks. The prior on the recurrent layer weights $\omega=\{W^x, W^h\}$ is a fully factorized standard normal distribution. The posterior is factorized over the rows of weights, and each factor is searched as a mixture of two normal distributions:
\begin{equation*}
q(w^x_k|m^x_k) = p^x \mathcal{N}(0, \sigma^2 I) + (1-p^x) \mathcal{N}(m^x_k, \sigma^2 I),\quad\:
\end{equation*}
\begin{equation}
q(w^h_j|m^h_j) = p^h \mathcal{N}(0, \sigma^2 I) + (1-p^h) \mathcal{N}(m^h_j, \sigma^2 I),
\end{equation}
\begin{equation*}
k=\overline{1,n}, \quad j=\overline{1,m}.\quad
\end{equation*}
%\begin{gather} 
Under assumption $\sigma\approx 0$ sampling the row of weights from such posterior means putting all the weights from this row either to 0 (drop the corresponding input neuron) or to some learned values. Thus this model is a probabilistic analog of binary dropout with dropout rates $p^x$ and $p^h$.

After unfolding the recurrence in the network, the maximization of the variational lower bound for such model looks as follows:
\begin{gather}
\sum_{i=1}^\ell \int q(\omega|M) \log \Bigl(y^i\big|f_y\bigl(f_h(x^i_T, f_h(\dots f_h (x^i_1, h^i_0))\bigr)\Bigr) d \omega - \notag \\
-KL\Bigl(q(\omega|M)\big\|p(\omega)\Bigr) \rightarrow \max_{M} 
\label{unfold}
\end{gather}

Each integral in the first part of~\eqref{unfold} is estimated with MC integration with a single sample $\hat \omega_i \sim q(\omega|M)$.
To make this estimation unbiased: (a) the weights sample $\hat \omega_i$ should remain the same for all time steps $t=\overline{1, T}$ for a fixed object; (b) dropout rates $p^x$ and $p^h$ should be fixed because the distribution we are sampling from depends on them.
The KL-divergence term from~\eqref{unfold} is approximately equivalent to  $L_2$ regularization of the variational parameters $M$.

Finally, this probabilistic model leads to the following dropout application in RNNs: we sample a binary mask for the input and hidden neurons, one mask per object for all moments of time, and optimize the $L_2$-regularized log-likelihood with the dropout rates and the weight of $L_2$-regularization chosen using cross-validation. Also, the same dropout technique may be applied to forward connections in RNNs, for example in embedding and dense layers~\cite{gal}. 

The same technique can be applied to more complex architectures like LSTM in which the information flow between input and hidden units is controlled by the gate elements:
%\begin{equation}
\begin{equation*}
\label{LSTM1}
i = sigm(h_{t-1}W^h_i + x_t W^x_i) \quad
o = sigm(h_{t-1}W^h_o + x_t W^x_o) 
\end{equation*}
\begin{equation*}
f = sigm(h_{t-1}W^h_f + x_t W^x_f) \quad 
%\end{equation}
%\begin{equation}
\label{LSTM2}
g = tanh(h_{t-1} W^h_g + x_t W^x_g) 
\end{equation*}
\begin{equation}
c_t = f \odot c_{t-1} +  i \odot  g \quad
h_t =  o \odot tanh(c_t)
\end{equation}%\end{equation}
Here binary dropout masks for input and hidden neurons are generated 4 times: individually for each of the gates $i,o,f$ and input modulation $g$. 

\section{Variational Dropout for RNN sparsification}
Dropout for RNNs proposed by Gal \& Ghahramani~\yrcite{gal} helps to avoid overfitting but is very sensitive to the choice of the dropout rates.
On the other hand, Sparse VD allows automatic tuning of the Gaussian dropout parameters individually for each weight which results in the model sparsification. We combine these two techniques to sparsify and regularize RNNs. 

Following Molchanov et al.~\yrcite{dmolch}, we use the fully factorized log-uniform prior and approximate the posterior with a fully factorized normal distribution over the weights $\omega=\{W^x, W^h\}$:
\begin{equation}
\begin{gathered}
q(w^x_{ki}|m^x_{ki}, \sigma^x_{ki}) = \mathcal{N}\bigl(m^x_{ki}, {\sigma^x_{ki}}^2\bigr), \:\\
q(w^h_{ji}|m^h_{ji}, \sigma^h_{ji}) = \mathcal{N}\bigl(m^h_{ji}, {\sigma^h_{ji}}^2\bigr), %\\
%\sigma^x_{ki}=\alpha^x_{ki} (m^x_{ki})^2, %\sigma^h_{ji}=\alpha^h_{ji} (m^h_{ji})^2. \\
\end{gathered}
\end{equation}
where $\sigma^x_{ki}$ and $\sigma^h_{ji}$ have the same meaning as in additive reparameterization~\eqref{ART}.

To train the model, we maximize the variational lower bound approximation 
\begin{gather} 
\sum_{i=1}^\ell \int q(\omega|M, \sigma) \log \Bigl(y^i\big|f_y\bigl(f_h(x^i_T, f_h(\dots f_h (x^i_1, h^i_0))\bigr)\Bigr) d \omega - \notag \\
- \sum_{k,i=1}^{n,m} k\biggl(\frac{{\sigma^x_{ki}}^2}{{m^x_{ki}}^2}\biggr) - \sum_{j,i=1}^{m,m} k\biggl(\frac{{\sigma^h_{ji}}^2}{{m^h_{ji}}^2}\biggr)
\label{VLBour}
\end{gather}

w.\,r.\,t. $\{M, \log \sigma\}$ using stochastic mini-batch methods. Here the recurrence in the expected log-likelihood term is unfolded as in~\eqref{unfold} and the KL is approximated using~\eqref{kl}. The integral in~\eqref{VLBour} is estimated with a single sample $\hat \omega_i \sim q(\omega|M, \alpha)$ per input sequence. 
We use the reparameterization trick (for unbiased integral estimation) and additive reparameterization (for gradients variance reduction) to sample both input-to-hidden and hidden-to-hidden weight matrices $\widehat W^x, \widehat W^h$. To reduce the variance of the gradients and for more computational efficiency we also apply the local reparameterization trick to input-to-hidden matrix $\widehat W^x$ moving the noise from the weights to the preactivations:
\begin{equation}
\begin{gathered}
(x_t \widehat W^x)_j = \sum_{k=1}^n x_{t,k} m^x_{kj}  + 
\epsilon_j \sqrt{\sum_{k=1}^n  x^2_{t,k} {\sigma^x_{kj}}^2}\:, \\
\epsilon_j \sim \mathcal{N}(0, 1).
\end{gathered}
\end{equation}
As a result, only 2-dimensional noise on input-to-hidden connections is required for each mini-batch: we generate one noise vector of length $m$ for each object in a mini-batch.

The local reparameterization trick cannot be applied to the hidden-to-hidden matrix $W^h$. We use the same sample $\widehat W^h$ for all moments of time, therefore in the multiplication $h_{t-1} \widehat W^h$ the vector $h_{t-1}$ depends on $\widehat W^h$ and the rule about the sum of normally distributed random variables cannot be applied. Since usage of 3-dimensional noise (2 dimensions of $\widehat W^h$ and a mini-batch size) is too resource-consuming we sample one noise matrix for all objects in a mini-batch for efficiency:
\begin{equation}
\hat w^h_{ji}=m^h_{ji}+\sigma^j_{ji}\epsilon^h_{ji},\quad \epsilon^h_{ji} \sim \mathcal{N}(0, 1).
\end{equation}

The final framework works as follows: we sample Gaussian additive noise on the input-to-hidden preactivations (one per input sequence) and hidden-to-hidden weight matrix (one per mini-batch), optimize the variational lower bound~\eqref{VLBour} w.\,r.\,t. 
$\{M, \log \sigma\}$, and for many weights we obtain the posterior in the form of a zero-centered $\delta$-function because the KL-divergence encourages sparsity. These weights can then be safely removed from the model.

In LSTM the same prior-posterior pair is consisered for all input-to-hidden and hidden-to-hidden matrices and all computations stay the same. The noise matrices for input-to-hidden and hidden-to-hidden connections are generated individually for each of the gates $i,o,f$ and input modulation $g$.

\section{Experiments}

We perform experiments with LSTM as the most popular recurrent architecture nowadays. We use Theano~\cite{theano} and Lasagne~\cite{lasagne} for implementation. The source code will be available soon at \url{https://github.com/tipt0p/SparseBayesianRNN}.  We demonstrate the effectiveness of our approach on two diverse problems: Character Level Language Modeling and Sentiment Analysis. Our results show that Sparse Variational Dropout leads to a high level of sparsity in recurrent models without a significant quality drop. 

We use the dropout technique of Gal \& Ghahramani~\yrcite{gal} as a baseline because it is the most similar dropout technique to our approach and denote it VBD (variational binary dropout). 

According to Molchanov et al.~\yrcite{dmolch}, training neural networks with Sparse Variational Dropout from a random initialization is troublesome, as a lot of weights may become pruned away before they could possibly learn something useful from the data. We observe the same effect in our experiments with LSTMs, especially with more complex models. LSTM trained from a random initialization may have high sparsity level, but also have a noticeable quality drop. To overcome this issue we start from pre-trained models that we obtain by training networks without Sparse Variational Dropout for several epochs. 

Weights in models with Sparse Variational Dropout cannot converge exactly to zero because of the stochastic nature of the training procedure. To obtain sparse networks we explicitly put weights with high corresponding dropout rates to 0 during testing as in Molchanov et al.~\yrcite{dmolch}. We use the value $\log\alpha = 3$ as a threshold. 
%This value corresponds to a Binary Dropout rate p > 0.95. 

For all weights that we sparsify using Sparse Variational Dropout, we initialize $\log{\sigma^2}$ with -6. 
We optimize our networks using Adam~\cite{adam}. 

Networks without any dropout overfit for both our tasks, therefore, we present results for them with early stopping. Throughout experiments we use the mean values of the weights to evaluate the model quality (we do not sample weights from posterior on the evaluating phase). This is a common practice when working with dropout.

\subsection{Sentiment Analysis}
\vspace{-1mm}
\paragraph{Data.} Following Gal \& Ghahramani \yrcite{gal} we evaluated our approach on the sentiment analysis regression task. The dataset is constructed based on Cornell film reviews corpus collected by Pang \& Lee \yrcite{regr_data}. It consists of approximately 10 thousands non-overlapping segments of 200 words from the reviews. The task is to predict corresponding film scores from 0 to 1. We use the provided train and test partitions.
\vspace{-2mm}
\paragraph{Setup.} We use networks with one embedding layer of 128 units, one LSTM layer of 128 hidden units, and finally, a fully connected layer applied to the last output of the LSTM (resulting in a scalar output). All weights are initialized in the same way as in Gal \& Ghahramani \yrcite{gal}. We train our networks using batches of size 128 and a learning rate of 0.001 for 1000 epochs. We also clip the gradients with threshold 0.1. For all layers with VBD we use dropout rate 0.3 and weight decay $10^{-3}$ (these parameters are chosen using cross validation).
\vspace{-2mm}
\paragraph{Results.}
\begin{table}[t]
\caption{Results on sentiment regression task. Prediction quality is reported in MSE (the lower the better). Sparsity levels reported for $W^x$ and $W^h$ separately in percents of zero weights. For Sparse VD methods initialization types are reported in brackets.}
\label{sentiment}
\vskip -0.15in
\begin{center}
\begin{small}
%\begin{sc}
\begin{tabular}{lcccr}
\hline
\abovespace\belowspace
Method (init) & Test MSE & Sparsity x~- h \% \\
\hline
\abovespace
No dropout & 0.1518 &  --\\
\belowspace
VBD & 0.1488 &  --\\
\hline
\abovespace
SparseVD (random)  & 0.1526&  99.91~-- 99.90\\
SparseVD (no dropout)  & 0.1503&  {\bf 99.95~-- 99.92}\\
\belowspace
SparseVD (VBD)  & {\bf 0.1475}& 99.63~-- 99.49\\
\hline
\end{tabular}
%\end{sc}
\end{small}
\end{center}
\vskip -0.15in
\end{table}
As a baseline, we train the network without any dropout and with VBD on all layers. 

In this experiment, our goal is to check the applicability of Sparse VD for recurrent networks, therefore we apply it only to LSTM layer. For embedding and dense layers we use VBD. 

We try both start training of the network with Sparse VD from random initialization and from two different pre-trained models. The first pre-trained model is obtained after 4 epochs of training of the network without any dropout. The second one is obtained after 200 epochs of training of the network with VBD on all layers. 
We choose number of pretraining epochs using models quality on cross-validation.

The results are shown in Table~\ref{sentiment}. In this task our approach achieves extremely high sparsity level both from random initialization and from pre-trained models. Sparse VD networks trained from pre-trained models achieve even better quality than baselines. Note that models already have this sparsity level after approximately 20 epochs. 

\begin{figure*}[h]
    \centering
        \centering
                \includegraphics[width=16cm]{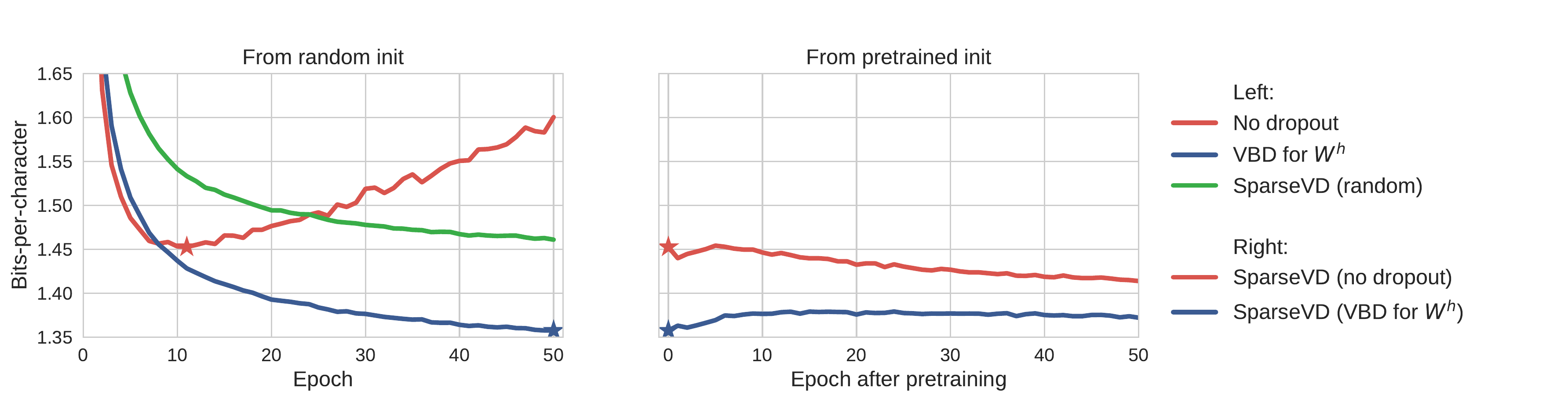}
        %\caption{Character level Language Modelling task: results on test set in bits-per-character (lower is better). Left~--- training from random initialization, right~--- training from pretrained initialization. Stars correspond to pretrained models.}
        
        \caption{Results on character level Language Modelling task: prediction quality on the test set in bits-per-character (lower is better). Left~--- training from random initialization, right~--- training from pretrained initialization. Stars correspond to pretrained models. For Sparse VD methods initialization types are reported in brackets.}
        
        \label{fig:acc}
\end{figure*}

\begin{figure}[h]
    \centering
        \centering
                \includegraphics[width=8.3cm]{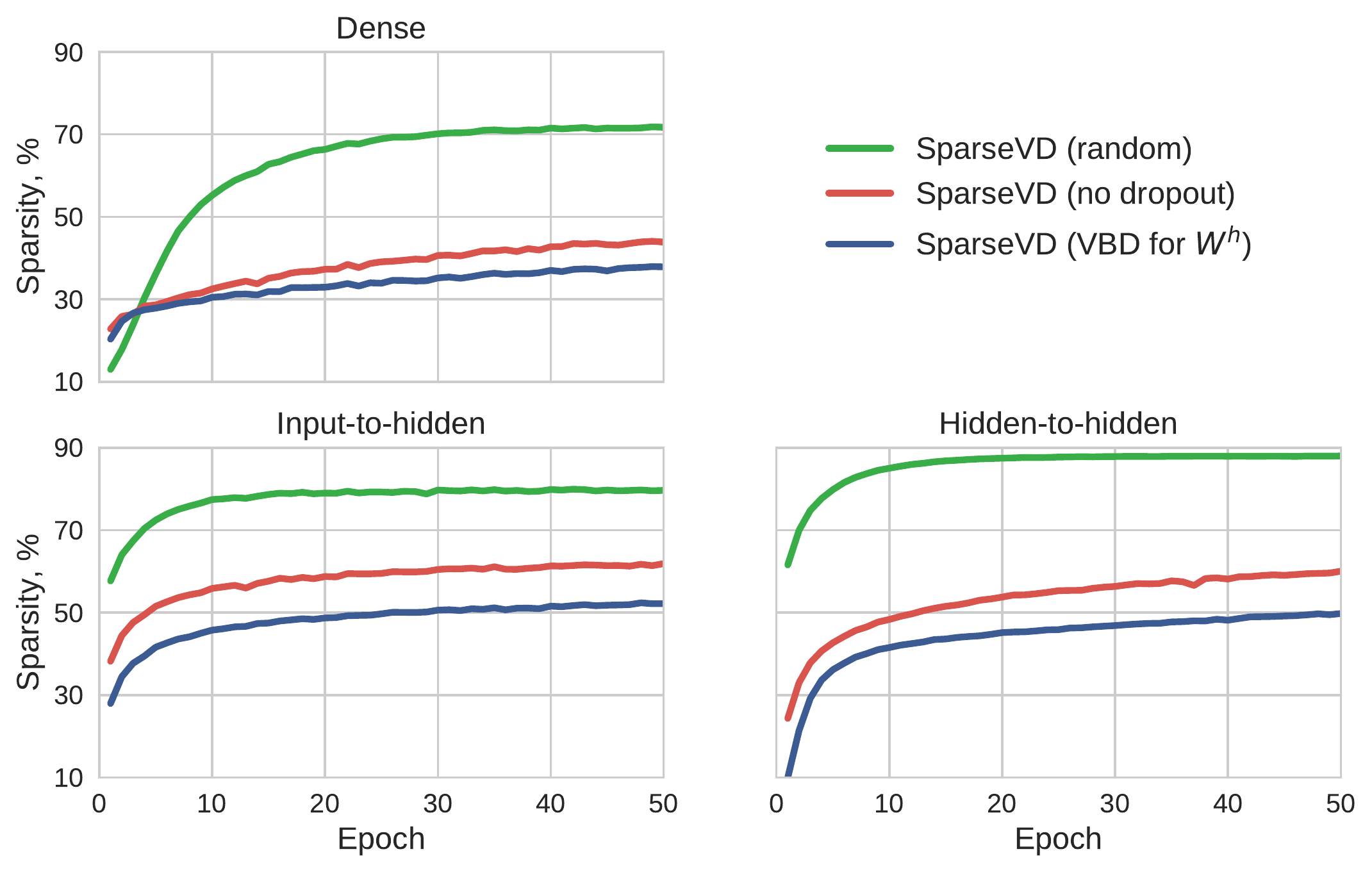}
        \caption{Results on character level Language Modelling task: sparsity levels for $W^y,W^x,W^h$ in LSTM. Initialization types are reported in brackets.}
        \label{fig:ard}
\end{figure}

\subsection{Character Level Language Modeling}
\vspace{-1mm}
\paragraph{Data.} Following Mikolov et al. \yrcite{mikolov11} we use the Penn Treebank Corpus to train our Language Model (LM). The dataset contains approximately 6 million characters and a vocabulary of 50 characters. We use the provided train, validation and test partitions.
\vspace{-2mm}
\paragraph{Setup.} We use networks with one LSTM layer of 1000 hidden units to solve the character level LM task. 
All weight matrices of the networks are initialized orthogonally and all biases are initialized with zeros. Initial values of hidden and cell elements are trainable and also initialized with zeros.
We train our networks on non-overlapping sequences of 100 characters in batches of 64 using a learning rate of 0.002 for 50 epochs, and clip gradients with threshold 1. For all layers with VBD we use dropout rate 0.25 and do not use weight decay (these parameters are chosen using quality of VDB model on validation set).
\vspace{-2mm}
\paragraph{Results.}

As a baseline, we train the network without any dropout and with VBD only on recurrent weights (hidden-to-hidden). Semeniuta et al.~\yrcite{semeniuta16} showed that for this particular task applying dropout for feed-forward connections additionally to VBD on recurrent ones does not improve the network quality. We observe the same effect in our experiments.

In this experiment we try to sparsify both LSTM and dense layers therefore we apply Sparse VD for all layers. We try both start training of the network with Sparse VD from random initialization and from two different pre-trained models. The first pre-trained model is obtained after 11 epochs of training of the network without any dropout. The second one is obtained after 50 epochs of training of the network with VBD on recurrent connections. We choose the number of pretraining epochs using models quality on validation set.

The results are shown in Table~\ref{charlm}. Here we do not achieve such extreme sparsity level as in the previous experiment. 
This effect may be a consequence 
of the higher complexity of the task. Also in LM problem we have several outputs for one input sequence (one output for each time step) instead of one output in Sentiment regression. As a result the log-likelihood part of the loss function is much stronger for LM task and regularizer can not sparsify the network so effectively.
Here we see that the balance between the likelihood and the regularizer varies a lot for different tasks with RNNs and should be explored futher. 

Fig.~\ref{fig:acc} and \ref{fig:ard} show the progress of test quality and network sparsity level through the training process. 
Sparse VD network trained from random initialization underfits and therefore has a slight quality drop in comparison to baseline network without regularization. 
Sparse VD networks trained from pre-trained models achieve much higher quality but have lower sparsity levels than the one trained from random initialization.
Better pretrained models are harder to sparsify. The quality of the model pretrained with VBD drops on the first epoches while the sparsity grows, and the model does not fully recover later.

\begin{table}[t]
\setlength{\tabcolsep}{4pt}
\caption{Results on character level Language Modelling task. Prediction quality is reported in bits-per-character (lower is better). Sparsity levels reported for $W^x$, $W^h$ and $W^y$ separately in percents of zero weights. For Sparse VD methods initialization types are reported in brackets.}
\label{charlm}
\vspace{-0.3cm}
%\vskip -0.7in
\begin{center}
\begin{small}
%\begin{sc}
\begin{tabular}{lcccr}
\hline
\abovespace\belowspace
Method (init) & Valid & Test & Sparsity x~- h~- y \% \\
\hline
\abovespace
No dropout    & 1.499 & 1.453 & -- \\
\belowspace
VBD for $W^h$ & {\bf 1.394} & {\bf 1.358} & --\\
\hline
\abovespace
SparseVD (random)  & 1.506& 1.461& {\bf 79.7~-- 88.0~-- 71.7}\\
SparseVD (no dropout)  & 1.454& 1.414& 61.9~-- 60.0~-- 43.9\\
\belowspace
SparseVD (VBD for $W^h$)  & 1.409 & 1.372 & 52.2~-- 49.8~-- 37.9\\
\hline
\end{tabular}
%\end{sc}
\end{small}
\end{center}
\vskip -0.15in
\end{table}

\section{Related Work}
\label{related}
\subsection{Regularization of RNNs}
Deep neural networks often suffer from overfitting, and different regularization techniques are used to improve their generalization ability. Dropout~\cite{dropout} is a popular method of neural networks regularization. The first successful implementations of this method for RNNs~\cite{pham13,zaremba14} applied dropout only for feed-forward connections and not recurrent ones. 

Introducing dropout in recurrent connections may lead to a better regularization technique but its straightforward implementation may results in underfitting and memory loss through time~\cite{semeniuta16}.  
Several ways of dropout application for recurrent connections in LSTM were proposed recently~\cite{moon15,gal,semeniuta16}. These methods inject binary noise into different parts of LSTM units. 
Semeniuta et al.~\yrcite{semeniuta16} shows that proper implementation of dropout for recurrent connections is important not only for effective regularization but also to avoid vanishing gradients.

Bayer et al.~\yrcite{bayer13} successfully applied fast dropout~\cite{fast_dropout}, a deterministic approximation of dropout, to RNNs. 
Krueger et al.~\yrcite{zoneout} introduced zoneout which forces some hidden units to maintain their previous values, like in feedforward stochastic depth networks~\cite{stoch}.

\subsection{Compression of RNNs}
Reducing RNN size is an important and rapidly developing area of research. One possible concept is to represent large RNN weight matrix by some approximation of the smaller size. For example, Tjandra et. al.~\yrcite{tjandra} use Tensor Train decomposition of the weight matrices and Le et al.~\yrcite{kroneker} approximate this matrix with Kronecker product. 
Hubara et. al~\yrcite{itay} limit the weights and activations to binary values proposing a way how to compute gradients w.\,r.\,t. them. 

Another concept is to start with a large network and to reduce its size during or after training. The most popular approach here is pruning: the weights of the RNN are cut off on some threshold. Narang et al.~\yrcite{pruning} choose threshold using several hyperparameters that control the frequency, the rate and the duration of the weights eliminating. 

\section{Discussion and future work}
When applying Sparse VD to RNNs we rely on the dropout for RNNs proposed by Gal \& Ghahramani~\yrcite{gal}. The reason is that this dropout technique for RNNs is the closest one to Sparse VD approach. However, there are several other dropout methods for recurrent networks that outperform this baseline \cite{semeniuta16,zoneout}. Comparison with them is our future work. Combining Sparse VD with these latest dropout recipes is also an interesting research direction. The challenge here is that the noise should be put on the neurons or gates instead of the weights as in our model. However, there are several recent papers~\cite{group, chris} where group sparsity methods are proposed for fully connected and convolutional networks. These methods can be used to solve the underlined problem.

The comparison of our approach with other RNN sparsification techniques is still a work-in-progress. 
It would be interesting to perform this comparison on larger networks, for example, for speech recognition task.

One more curious direction of the research is to sparsify not only recurrent layer but an embedding layer too. It may have a lot of parameters in the tasks with large dictionary, such as word based language modeling. 

\vspace{-0.3cm}
\section*{Acknowledgements} 
\vspace{-0.1cm}
We would like to thank Dmitry Molchanov and Arsenii Ashukha for valuable feedback.
Nadezhda Chirkova has been supported by Russian Academic Excellence Project `5-100', and Ekaterina Lobacheva has been supported by Russian Science Foundation grant 17-71-20072.
We would also like to thank the Department of Algorithms and Theory
of Programming, Faculty of Innovation and High Technology
in Moscow Institute of Physics and Technology for provided computational resources.

\bibliographystyle{icml2017}

\end{document}